\documentclass[11pt,a4paper]{article}
\usepackage{times,latexsym}
\usepackage{url}
\usepackage[T1]{fontenc}
\usepackage{amsmath}
\usepackage{graphicx}
\usepackage{booktabs}
\usepackage{xcolor}
\usepackage[frozencache=true,cachedir=mintedcache]{minted} 
\usepackage{tabularx}
\usepackage{multirow}
\usepackage{longtable}
\usepackage{listings}
\usepackage{caption}
\usepackage[acceptedWithA]{tacl2021v1}

\usepackage{xspace,mfirstuc,tabulary}

\usepackage{xcolor}

\newif\iftaclinstructions
\taclinstructionsfalse 
\iftaclinstructions
\renewcommand{\confidential}{}
\renewcommand{\anonsubtext}{(No author info supplied here, for consistency with
TACL-submission anonymization requirements)}
\newcommand{\instr}
\fi

\iftaclpubformat 

\else

\fi

\title{End-to-End Long Document Summarization using Gradient Caching}

\author{{Rohit Saxena \qquad Hao Tang \qquad Frank Keller} \\
  Institute for Language, Cognition and Computation \\
  School of Informatics, University of Edinburgh \\
  10 Crichton Street, Edinburgh EH8 9AB\\
  \texttt{rohit.saxena@ed.ac.uk}\quad \texttt{hao.tang@ed.ac.uk} \quad \texttt{keller@inf.ed.ac.uk}}

\date{}

\begin{document}

\maketitle

\begin{abstract}
Training transformer-based encoder-decoder models for long document summarization poses a significant challenge due to the quadratic memory consumption during training.
Several approaches have been proposed to extend the input length at test time, but training with these approaches is still difficult, requiring truncation of input documents and causing a mismatch between training and test conditions.
In this work, we propose CachED (Gradient \textbf{Cach}ing for \textbf{E}ncoder-\textbf{D}ecoder models), an approach that enables end-to-end training of existing transformer-based encoder-decoder models, using the entire document without truncation.
Specifically, we apply non-overlapping sliding windows to input documents, followed by fusion in decoder. 
During backpropagation, the gradients are cached at the decoder and are passed through the encoder in chunks by re-computing the hidden vectors, similar to gradient checkpointing.
In the experiments on long document summarization, we extend BART to CachED BART, processing more than 500K tokens during training and achieving superior performance without using any additional parameters.
\end{abstract}

\section{Introduction}

Summarization is a critical task in natural language understanding, aiming to reduce extensive information into its most essential content by generating a concise and coherent summary. In recent years, transformer-based~\citep{vaswani} pretrained language models have shown remarkable success in abstractive summarization, primarily on short texts~\citep{narayan-etal-2018-dont,nallapati-etal-2016-abstractive,gliwa-etal-2019-samsum}, heavily relying on dependencies within the input text or context of words. 

Despite their success, these models face significant challenges when applied to long document summarization tasks~\citep{shaham-etal-2022-scrolls,gorinski-lapata-2015-movie,kryscinski-etal-2022-booksum}. One of the primary limitations is their inability to handle long input during training, due to the memory requirement being quadratic with respect to the sequence length. This often necessitates truncation of the input text during training, resulting in a loss of crucial information and hampering the quality of the generated summaries. This problem is particularly pronounced in domains that require processing of extremely long text, such as book summarization~\citep{kryscinski-etal-2022-booksum}, where maintaining the full context is essential for producing accurate and meaningful summaries.

Prior work has attempted to address the limitation of processing long input, including designing attention mechanisms that are more memory efficient~\citep{Beltagy2020Longformer}, dividing an input document into chunks~\citep{unlimiformer, ivgi-etal-2023-efficient, xie2024chunkalignselectsimple, yen2024longcontextlanguagemodelingparallel}, or extending context at test time~\citep{ratner-etal-2023-parallel,han-etal-2024-lm-Infinite}.
Despite all the effort, truncation during training (typically at 16K tokens) is ubiquitous and is the standard approach to dealing with memory issues during training, causing a mismatch between training and test conditions.

To tackle the problem of truncation, we propose CachED (Gradient \textbf{Cach}ing for \textbf{E}ncoder-\textbf{D}ecoder Models), a simple and efficient approach that enables end-to-end training of existing encoder-decoder transformer models for long document summarization.
We follow the chunking approach in favor of its generality, allowing us to plug and play any pretrained models, but more importantly, providing us the opportunity to release memory between the computation of chunks.
We only keep the final output of the encoder, and release the intermediate results of the encoder whenever possible.
The fusion of encoder output happens at the decoder~\citep{izacard-grave-2021-leveraging}.
Gradients are computed as usual but are cached at the encoder output.
The cached gradients are then propagated to the encoder chunk by chunk.
The peak memory usage of our approach is greatly reduced, allowing us to train encoder-decoder models on entire input documents without truncation.

We apply our approach to BART (named CachED BART) on several long document summarization benchmarks, including GovReport, SummScreenFD, QMSum, ScriptBase, and BookSum. CachED BART achieves superior performance compared to existing approaches even when using a small model with a context size of 1024 tokens.
Our approach is also general and can be applied to any pretrained encoder-decoder models.

In summary, the contributions of this work are:
\begin{enumerate}
    \item We propose CachED\footnote{Our code is available at \href{https://github.com/saxenarohit/CachED}{github.com/saxenarohit/CachED}}, a simple and efficient approach that enables end-to-end training of any existing encoder-decoder transformer models on long input without truncation.
    \item We show that CachED BART achieves superior performance on extremely long document summarization, such as book summarization.
    \item Our results properly and correctly doing gradient descent without truncation can lead to improvements and strong performance.
\end{enumerate}

\section{Abstractive Summarization with Encoder-Decoder Models}

The task of summarization is to produce a summary of $M$ tokens $y_1, y_2, \dots, y_M$ given an input document of $L$ tokens $x_1, x_2, \dots, x_L$, where a token can be a word or a wordpiece~\citep{wu2016googlesneuralmachinetranslation}.
The dominant approach to abstractive summarization is to use an encoder-decoder model~\citep{bahdanau2016neuralmachinetranslationjointly,T5,Beltagy2020Longformer}, where the encoder turns the input document into a sequence of hidden vectors and the decoder produces a summary attending to the hidden vectors.
More formally,
\begin{align}
h_1, \dots, h_L = \text{Enc}(x_1, \dots, x_L),
\end{align}
where $\text{Enc}$ is the encoder, and
\begin{align}
y_m &= \text{Dec}(y_1, \dots, y_{m-1}, h_1, \dots, h_L) 
\end{align}
where the decoder $\text{Dec}$ is repeatedly called for $m=1, \dots, M$.
A transformer-based encoder typically consists layers of self-attention, and a vanilla implementation requires $O(L^2)$ of memory~\citep{vaswani}.
Long-document summarization is the setting where $L$ is large, making it difficult to store the intermediate results of the entire input in memory.

A naive approach to solving the memory problem is truncating the input, only taking, say, the first 16,000 tokens as input and capping the length at $\min(16,000, L)$.
Depending on the types of summarization, this approach can be sufficient, for example, for summarizing news articles.
For long documents, such as books~\citep{kryscinski-etal-2022-booksum} or movie scripts~\citep{saxena-keller-2024-select}, naively truncating the input makes it impossible to properly perform the task, as a model has no access to the truncated input.
Despite the obvious limitation, truncation is widely used during training \citep{Beltagy2020Longformer,guo-etal-2022-longt5,unlimiformer,xie2024chunkalignselectsimple}, and is sometimes the only option when scaling up the model size.

Another approach is to divide the input document into chunks, with each chunk encoded individually.
More formally, the input document of length $L$ is divided into $K$ chunks, with each chunk of size $\lfloor L / K \rfloor$. Each chunk, denoted as $x_{(k-1) \lfloor L/K \rfloor + 1}, \dots, x_{k \lfloor L/K \rfloor}$, is encoded into $h_{(k-1) \lfloor L/K \rfloor + 1}, \dots, h_{k \lfloor L/K \rfloor}$, for $k = 1, \dots, K$.
The memory requirement of this approach is $O(\lfloor L/K \rfloor^2)$ per chunk, i.e., $O(K \cdot \lfloor L/K \rfloor^2) = O(L^2 / K)$ in total, less than the $O(L^2)$ when running self-attention on the entire sequence.\footnote{The runtime complexity of the decoder is $O(M^2 + LM)$, and does not dominate $O(L^2)$ when $M < L$.} 
Diving the input document into chunks is sometimes called a chunk-based approach~\citep{xie2024chunkalignselectsimple}, the sliding window approach \cite{ivgi-etal-2023-efficient}, or parallel context~\citep{yen2024longcontextlanguagemodelingparallel,ratner-etal-2023-parallel}, in which chunks might or might not have overlaps.
This approach also makes a modeling assumption: text representation can only be contextualized within each chunk, delaying further contextualization or fusion in the decoder.

Despite the memory saving with sliding windows, the input documents are still truncated before chunking \citep{ivgi-etal-2023-efficient,unlimiformer,xie2024chunkalignselectsimple}, because intermediate results are not released from memory after the computation of each chunk.
Truncation of input documents leads to a mismatch between training and test conditions.
It is still an open question whether properly and correctly doing gradient descent end to end without truncation would be better than that with truncation, a question to be addressed in this paper.

\begin{figure*}
  \centering
  \includegraphics[width={\textwidth}]{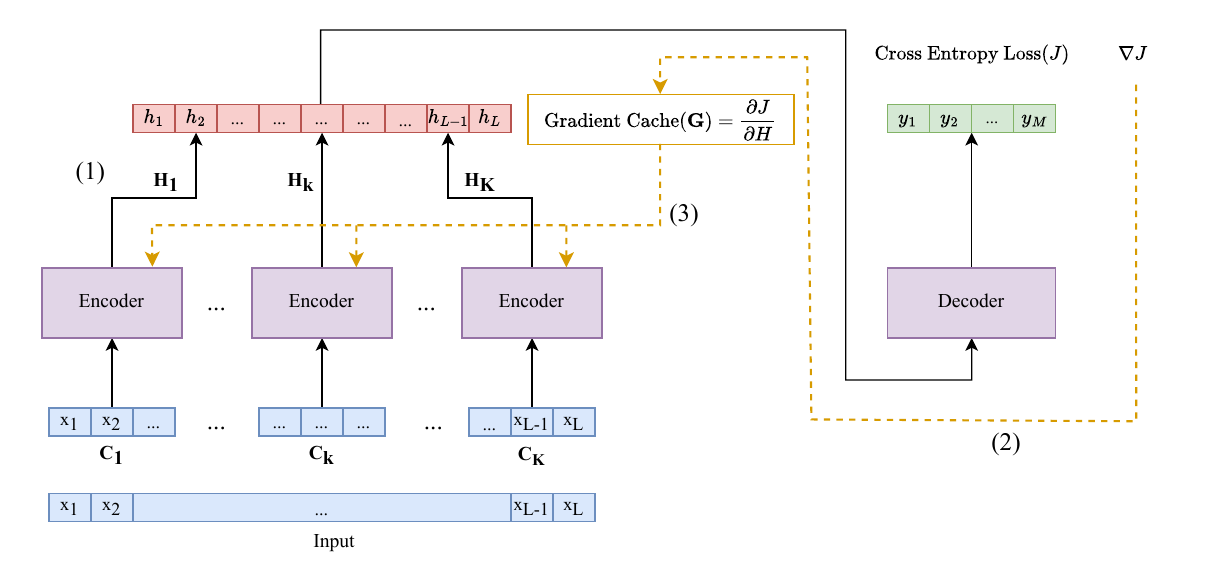}
  \caption{An overview of a CachED approach to long document summarization. The model is trained end-to-end by splitting and encoding the input into chunks (Step 1), then passing the concatenated hidden vectors to the decoder. Gradients are cached for the encoded tokens (Step 2) and passed back to the chunks individually (Step 3).}
  \label{fig:model}
\end{figure*}

\section{CachED: Gradient Caching for Encoder-Decoder Models}

To address the compromise of truncating input documents, we propose gradient caching for encoder-decoder models (CachED), an approach that turns any existing transformer-based encoder-decoder models into a model for long document summarization without truncation.

Recall that an input document of length $L$ is divided into $K$ chunks and fitting $O(L^2/K)$ in memory is still difficult, especially when there are many layers.
The $O(L^2/K)$ memory requirement is due to $K$ calls to the encoder, each of which requires \color{black}$O(L^2/K^2)$\color{black}.
Instead of maintaining $K$ calls simultaneously in memory, we decide to call the encoder $K$ times in sequence, releasing the memory after each call and reducing the memory requirement to \color{black}$O(L^2/K^2)$\color{black}.
This can be easily done during inference, but not maintaining all $K$ calls in memory makes computing the gradient difficult during training.

To compute the full gradient without truncation, we ideally want to break the computation up with respect to the $K$ chunks.
If we use $J$ to denote the loss function and $\Theta$ to denote the parameters in the encoder, the relationship between the derivative of the individual $K$ chunks and the gradient is 
\begin{align}
\frac{\partial J}{\partial \Theta} = \sum_{k=1}^K \frac{\partial J}{\partial H_k} \frac{\partial H_k}{\partial \Theta},
\end{align}
where $H_k = \begin{bmatrix} h_{(k-1) \lfloor L/K \rfloor + 1} & \cdots & h_{k \lfloor L/K \rfloor} \end{bmatrix}$ is the concatenation of the hidden vectors from the $k$-th chunk.
The total derivative naturally leads to the following three steps.
\begin{enumerate}
\item Compute the encoder output $H_k$ for $k=1, \dots, K$ in sequence without storing the intermediate layers.
\item Compute the loss $J$ based on $H_k$ and the gradient $\partial J / \partial H_k$ with regular backpropagation.
\item Re-compute $H_k$ and the intermediate layers and use the cached $\partial J / \partial H_k$ to compute $\partial H_k / \partial \Theta$ in sequence for $k = 1, \dots, K$, accumulating the final gradient to the parameters $\partial J / \partial \Theta$.
\end{enumerate}
Figure~\ref{fig:model} illustrates the process of the CachED approach.
Step 1 can be implemented as a simple for loop,
\begin{minted}[fontsize=\footnotesize]{python}
Hs = []
for k in range(K):
    Ck = X[k*(L//K):(k+1)*(L//K)]
    Hk = Enc(Ck)
    Hk.detach()
    Hs.append(Hk)
\end{minted}
where \texttt{X} is the concatenation of the tokens $x_1, \dots, x_L$ in the input document, and \texttt{Enc} is the forward function of the encoder, and \texttt{L // K} is the chunk size $\lfloor L / K \rfloor$.
Note that \texttt{detach()} makes it explicit that the intermediate results before \texttt{Hk} can be discarded and do not occupy memory.\footnote{We use the language of \texttt{pytorch}, such as \texttt{detach()}, to describe the implementation, but similar concepts exist in other automatic differentiation toolkits.}
Though we present this step as a \texttt{for} loop, the $K$ calls within the \texttt{for} loop are trivially parallelizable and can be batched.

Step 2 can be implemented with regular backpropagation as follows.
\begin{minted}[fontsize=\footnotesize]{python}
H = torch.cat(Hs)
H.retain_grad()
Yhat = Dec(Y, H)
loss = cross_entropy(Yhat, Y)
loss.backward()
\end{minted}
where \texttt{Dec} is the forward function of the decoder, and \texttt{Y} is the concatenation of output tokens $y_1, \dots, y_M$.
Since the gradients are not normally stored unless the variables are parameters, the call \texttt{retain\_grad()} is necessary to guarantee that the gradient to \texttt{H} is computed and cached when \texttt{loss.backward()} is called.

Step 3 continues the incomplete backpropagation from Step 2 to the encoder,
\begin{minted}[fontsize=\footnotesize]{python}
for k in range(K):
    Ck = X[k*(L//K):(k+1)*(L//K)]
    Hk = Enc(Ck)
    Gk = H.grad[k*(L//K):(k+1)*(L//K)] 
    torch.autograd.backward(Hk, Gk)
\end{minted}
Again, though this step is presented as a \texttt{for} loop, the $K$ calls within the \texttt{for} loop are trivially parallelizable and can be batched.
At the end of Step 3, we have the full gradient with respect to both the encoder and the decoder model parameters, and are ready to make a gradient update.

The CachED approach is reminiscent to gradient checkpointing~\citep{chen2016trainingdeepnetssublinear}.
However, our approach does not require low-level custom implementations, and as shown above, is applicable to any encoder-decoder models.
One drawback of the CachED approach is that the encoders need to be called twice, but the runtime cost is typically marginal if the $K$ calls are properly parallelized and batched.
Similar to \citet{ivgi-etal-2023-efficient}, ours is a fusion-in-decoder approach \citep{izacard-grave-2021-leveraging}.
We assume that the encoder can sufficiently contextualize input tokens within a chunk, while the decoder is responsible for managing long-range dependencies.
Similar to fusion in decoder, we do not modify the positional encoding of the underlying model, making our method independent of the positional embedding used by the backbone.
We will study the runtime, memory, and efficacy of our approach in the experiments.

\section{Experimental Settings}

To showcase the CachED approach, we use BART and T5 as the backbone models. BART is chosen because it performs well on short text summarization but is less effective on longer text due to its input size limit, while T5 offers robust performance across diverse tasks. We will show how applying our approach to BART and T5, resulting in CachED BART and CachED T5, significantly outperforms their respective baselines, highlighting the benefits of our method. We experiment with both BART$_{\text{base}}$ and BART$_{\text{large}}$, as well as T5$_{\text{large}}$, comparing to other approaches that also fine-tune these models. In all experiments, we use a chunk size of 1024 tokens for BART and a context size of 512 tokens for T5. See Appendix~\ref{apn:hparams} for more details on the implementation and hyperparameters of the model.

We report ROUGE F1 (1/2/L) scores \citep{lin-2004-rouge} and BERTScore F1 \citep{bert-score} to evaluate the performance of our method on long document summarization tasks.

\begin{figure}
  \centering
  \includegraphics[width=0.9\columnwidth]{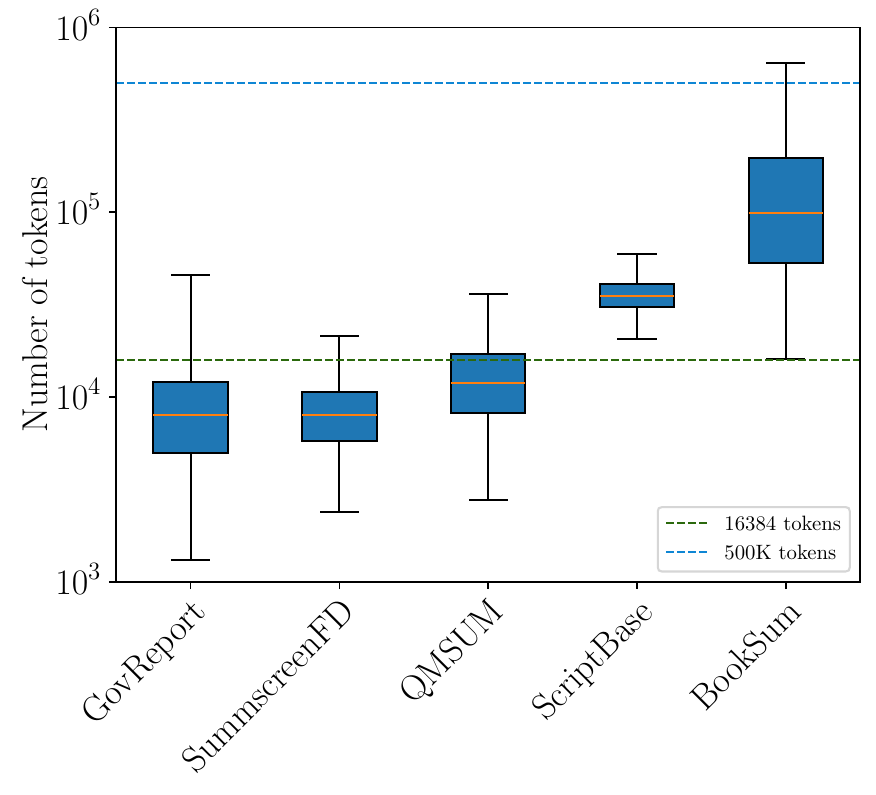}
  \caption{The mean and standard deviation of document lengths across summarization datasets, plotted on log scale.}
  \label{fig:length_dist}
\end{figure}

\subsection{Datasets}
We categorize the long document summarization datasets into 1) long documents, with a mean token length of less than 16K tokens, and 2) extremely long documents, with a mean length greater than 16K tokens. Figure~\ref{fig:length_dist} shows the mean and standard deviation of the input document length in tokens for the datasets we used.

\subsubsection{Long Document Summarization}

\noindent\textbf{GovReport} \citep{huang-etal-2021-efficient} is a large-scale summarization dataset consisting of reports published by the U.S.\ Government Accountability Office on national policy issues. The task is to write an executive summary of each report. The mean length of the documents is 9,616 tokens. 

\noindent\textbf{SummScreenFD} \citep{chen-etal-2022-summscreen} consists of community-contributed transcripts of television show episodes collected from the ForeverDream (FD) website. The summaries are recaps collected from Wikipedia and TVMaze. The mean length of the documents is 8,417 tokens.

\noindent\textbf{QMSUM} \citep{zhong-etal-2021-qmsum} is a query-based multi-domain meeting summarization dataset. The dataset consists of tuples of a query, document, and its corresponding summary. The mean length of the documents is 13,325 tokens.

\subsubsection{Extremely Long Document Summarization}

\textbf{ScriptBase} \citep{gorinski-lapata-2015-movie} consists of full-length movie scripts and corresponding Wikipedia summaries. We use the dataset splits from \citet{saxena-keller-2024-select}. The mean length of the movie scripts is 35,956 tokens.

\noindent\textbf{BookSum} \citep{kryscinski-etal-2022-booksum} consists of book text along with their summaries. BookSum features three subsets: paragraph, chapter, and book-level tasks. We focus on the most challenging BOOKSUM-Book task, which involves generating a summary of an entire book using the full text of the book. The mean length of the books is 139,219 tokens. The longest document in this dataset consists of 642,376 number of tokens.

\subsection{Baselines}

\noindent\textbf{BART}~\citep{lewis-etal-2020-bart} is a pretrained encoder-decoder model with 139M parameters in  BART$_{\text{base}}$ and 406M parameters in BART$_{\text{large}}$. Its maximum input sequence length is 1024 tokens. We fine-tune both the base model and the large model each dataset separately. 

\color{black}
\noindent\textbf{T5}~\citep{T5} is a pretrained encoder-decoder model designed for text-to-text tasks. We fine-tune T5$_{\text{large}}$  which contains 770M parameters. Its maximum input length is 512 tokens.
\color{black}
\noindent\textbf{LED}~\citep{Beltagy2020Longformer} is a Longformer-Encoder-Decoder (162M) parameters with a maximum input length of 16,384 tokens. We fine-tune the base version of this model.

\noindent\textbf{Unlimiformer}~\citep{unlimiformer} augments pretrained encoder-decoders and offloads
the cross-attention computation to a $k$NN index, allowing for unlimited context. We compare our approach with their custom fine-tuned BART$_{\text{base}}$ and PRIMERA model as its backbone. The maximum number of tokens is 16K during training. For PRIMERA, we report numbers from the paper as we could not replicate the result.

\noindent\textbf{SLED}~\citep{ivgi-etal-2023-efficient} extends pretrained encoder-decoder models for longer contexts by encoding the long input in chunks, and applying fusion in decoder.
Since ours and theirs are both plug-and-play approaches, we compare our approach with theirs applied to BART$_{\text{base}}$ and BART$_{\text{large}}$ (their best settings). During training, the maximum context length is 16K.

\begin{table*}[t!]
\caption{Test results on long document summarization datasets using different base models.
The best metric in every dataset is marked in \textbf{bold}.
Some results of Unlimiformer with PRIMERA are not reported due to out-of-memory issues.} 
\label{tab:long_doc_sum}
\vspace{1em}
\centering
\scalebox{0.85}{%
\begin{tabular}{lllll}
\toprule
Method & Parameters & \multicolumn{3}{c}{ROUGE 1 / 2 / L / BERTScore F1} \\  
& & \multicolumn{1}{c}{\textbf{GovReport}} & \multicolumn{1}{c}{\textbf{SummScreenFD}} &\multicolumn{1}{c}{\textbf{QMSum}}  \\
\midrule
\color{black} SFT (T5$_{\text{large}}$) & \color{black} 770M & \color{black} 45.3 / 18.8 / 22.4 / 61.7 & \color{black} 28.9 / 5.3 / 16.9 / 58.1 & \color{black} 29.6 / 7.0 / 19.4 / 57.6 \\
SFT (BART$_{\text{base}}$) & 139M & 50.2 / 19.0 / 23.7 / 63.4 & 31.5 / 6.7 / 18.1 / 58.2 & 32.9 / 8.8 / 21.1 / 59.5 \\
SFT (BART$_{\text{large}}$) & 406M & 54.4 / 21.1 / 25.1 / 65.6 & 34.1 / 7.5 / 18.9 / 59.6 & 33.0 / 7.9 / 20.0 / 59.5 \\
SFT (LED$_{\text{base}}$) & 162M & 56.3 / 25.8 / 27.4 / 65.6 & 33.8 / 8.3 / 19.8 / 59.5 & 30.9 / 7.8 / 19.5 / 57.6 \\
\midrule
SLED (BART$_{\text{base}}$) & 139M & 54.7 / 24.4 / 25.4 / 67.0 & 32.7 / 7.9 / 19.1 / 58.4 & 33.8 / 11.7 / 22.6 / 59.1 \\
Unlim. (BART$_{\text{base}}$) & 139M & 56.6 / \textbf{26.3} / 27.6 / \textbf{68.2} &  34.7 / 8.5 / 19.9 / 58.5 & 30.9 / 8.00 / 19.9 / 58.2  \\ 
SLED (BART$_{\text{large}}$) & 406M & \textbf{57.5} / \textbf{26.3} / 27.4 / 66.9 & 35.2 / 8.7 / 19.4 / 59.9 & 34.2 / 11.0 / 22.0 / 58.3 \\
Unlim. (PRIMERA) & 447M & 57.4 / 26.2 / 28.0 / 68.1 & 33.3 / 7.6 / 18.9 / 57.7 & \\
 \midrule
\color{black}
CachED T5$_{\text{large}}$ & \color{black}770M & \color{black} 51.9 / 22.5 / 24.5 / 63.4 & \color{black} 32.8 / 8.2 / 19.6 / 60.2 & \color{black} 32.9 / 8.4 / 19.9 / 59.4 \\
CachED BART$_{\text{base}}$ & 139M & 56.8 / \textbf{26.3} / 27.8 / 67.0 & 36.6 / 8.8 / 19.9 / 61.3 & 38.4 / 13.5 / 24.4 / 62.2 \\
CachED BART$_{\text{large}}$ & 406M & 57.0 / \textbf{26.3} / \textbf{28.19} / 67.3 & \textbf{37.2} / \textbf{9.1} / \textbf{20.1} / \textbf{61.59} & \textbf{38.9} / \textbf{14.0} / \textbf{24.6} / \textbf{62.5} \\
\bottomrule
\end{tabular} 
}
\end{table*}

\begin{table*}[t!]
\caption{Test results on extremely long document summarization datasets using different base models.
The best metric in every dataset is marked in \textbf{bold}.
Some results of Unlimiformer with PRIMERA are not reported due to out-of-memory issues.}
\label{tab:extreme_sum}
\vspace{1em}
\centering
\scalebox{0.95}{
\begin{tabular}{llll}
\toprule
Method & & \multicolumn{2}{c}{ROUGE 1 / 2 / L / BERTScore F1} \\
& & \textbf{ScriptBase} & \textbf{BookSum} \\
\midrule
\color{black} SFT (T5$_{\text{large}}$)& &\color{black} 33.4 / 4.5 / 14.5 / 55.9  & \color{black} 19.9 / 3.0 / 11.4 / 47.2  \\
SFT (BART$_{\text{base}}$)& & 39.2 / 8.2 / 17.2 / 57.8  & 23.6 / 5.0 / 13.2 / 49.0  \\
SFT (BART$_{\text{large}}$)& & 42.9 / 8.9 / 17.8 / 60.1  & 24.7 / 5.8 / 14.0 / 48.3 \\
SFT (LED$_{\text{base}}$) & & 45.5 / 10.5 / 19.4 / 60.8 & 26.2 / 3.8 / 16.9 / 47.3 \\
\midrule
Unlim. (PRIMERA) & &  & 38.2 / 7.1 / 16.0 / \\
Unlim. (BART$_{\text{base}}$) & & 44.8 / 12.3 / 18.3 / 58.7  & 36.7 / 7.3 / 15.5 / 51.5 \\ 
SLED (BART$_{\text{large}}$) & & 45.2 / 11.9 / 17.8 / 58.3 & 38.9 / 7.5 / 15.8 / 52.4  \\
 \midrule
\color{black} CachED T5$_{\text{large}}$ & & \color{black} 36.7 / 7.7 / 17.6 / 56.9 & \color{black} 29.5 / 5.6  / 15.9  / 49.8 \\
CachED BART$_{\text{base}}$ & & 48.9 / 14.4 / 19.8 / 64.1 & 39.4 / 9.2  / 17.0  / 53.6 \\
CachED BART${_\text{large}}$ & & \textbf{50.2} / \textbf{14.9} / \textbf{20.4} 
 / \textbf{64.6} & \textbf{42.8} / \textbf{10.5} / \textbf{18.8} / \textbf{54.4} \\
\bottomrule
\end{tabular}
}
\end{table*}

\section{Results}

Since GovReport, SummScreenFD, QMSum, and BookSum are commonly used benchmark, we quote the ROUGE (1/2/L) results from previous studies and adding BERTScore by running the baseline models ourselves.\footnote{We do see ROUGE (1/2/L) of the baseline models in the ballpark of those reported from previous studies, successfully replicating them.}
The results of the baseline models for ScriptBase, however, are never reported and are run by ourselves.

\subsection{Long Document Summarization}

Table \ref{tab:long_doc_sum} presents the evaluation results of various models on the long document summarization datasets: GovReport, SummScreenFD, and QMSum.
Our CachED approach demonstrates superior performance on the SummScreenFD and QMSum datasets and competitive performance on the GovReport dataset compared to the baseline models and previously proposed methods. Both CachED BART$_{\text{base}}$ and CachED BART$_{\text{large}}$ achieve substantial improvements over standard fine-tuning (SFT) and other competitive approaches such as SLED and Unlimiformer. CachED BART$_{\text{base}}$ outperforms bigger models on SummScreenFD and QMSum dataset highlighting the effectiveness of our method in processing long context. \color{black} In all the experiments, CachED T5$_{\text{large}}$, did not surpass CachED BART but demonstrated a notable improvement over standard T5 fine-tuning, showcasing the applicability of our method across different backbone architectures. \color{black}

\paragraph{GovReport} For the GovReport dataset, CachED BART$_{\text{base}}$ achieves a ROUGE-2 score of 26.3, matching the best performance among all models. The large version, CachED BART$_{\text{large}}$, further improves the ROUGE-L score to 28.19 compared to the previous methods. SLED (BART$_{\text{large}}$) achieves slightly better ROUGE-1 across the models, and Unlimiformer (BART$_{\text{base}}$) achieves the best BERTScore F1.

\paragraph{SummScreenFD} On the SummScreenFD dataset, CachED BART$_{\text{large}}$ outperforms all the models with ROUGE scores of 37.2/9.1/20.1. These results surpass all competing models, including Unlimiformer (PRIMERA) and SLED (BART$_{\text{large}}$) on ROUGE and BERTScore.

\paragraph{QMSum} For the QMSum dataset, CachED BART$_{\text{large}}$ achieves the highest scores across all metrics with a ROUGE 38.9/14.0/24.6. Unlimiformer performs poorly, even worse compared to standard fine-tuned BART with a 1024-token context. Compared to SLED, our approach is better by 5.7/3.0/2.6 ROUGE scores, a substantial advantage without even using additional parameters.

\subsection{Extremely Long Document Summarization}

Table \ref{tab:extreme_sum} presents the evaluation results of various models on ScriptBase and BookSum, where we report ROUGE (1/2/L) and BERTScore F1 metrics. Our proposed method substantially outperforms the baseline and previous methods across both datasets, showcasing its efficacy in handling extremely long document summarization tasks.

\paragraph{ScriptBase} For the ScriptBase dataset, CachED BART$_{\text{base}}$ achieves a notable improvement with ROUGE-1 of 48.9, ROUGE-2 of 14.4, ROUGE-L of 19.8, and BERTScore F1 of 64.1. The large version, CachED BART$_{\text{large}}$, further enhances performance, setting new state-of-the-art scores with ROUGE-1 of 50.2, ROUGE-2 of 14.9, ROUGE-L of 20.4, and BERTScore F1 of 64.6.

\paragraph{BookSum} On the BookSum dataset, CachED BART$_{\text{base}}$ demonstrates substantial gains with ROUGE scores of 39.4/9.2/17.0, and BERTScore F1 of 53.6. The large version, CachED BART$_{\text{large}}$, achieves the highest scores across all metrics, improving ROUGE 1/2/L scores by 3.9/3/3 compared to the best previous method. Appendix~\ref{sec:generatedsumm} shows generated summary of a book using our method.

\subsection{Summary}

In both dataset categories, CachED BART$_{\text{base}}$ shows substantial improvements over other variants of BART, especially on extremely long documents, compared to models twice its size in terms of the number of parameters. \color{black}Similar to long document datasets, CachED T5$_{\text{large}}$ did not surpass CachED BART but demonstrated considerable improvement over T5$_{\text{large}}$ standard fine-tuning on extremely long document datasets.\color{black} 

Similar to SLED, our approach is plug-and-play, applicable to any existing encoder-decoder models.
Comparing to SLED, the fact that our approach performs better shows that properly doing gradient descent without truncation can significantly improve performance.
Together with SLED, our results show that a context as small as 1024 tokens can have strong performance compared to models with much longer context. 
Overall, CachED BART$_{\text{large}}$ outperforms CachED BART$_{\text{base}}$, showcasing that our approach can also benefit from scale.
\begin{figure}
  \centering
  \includegraphics[width=\columnwidth]{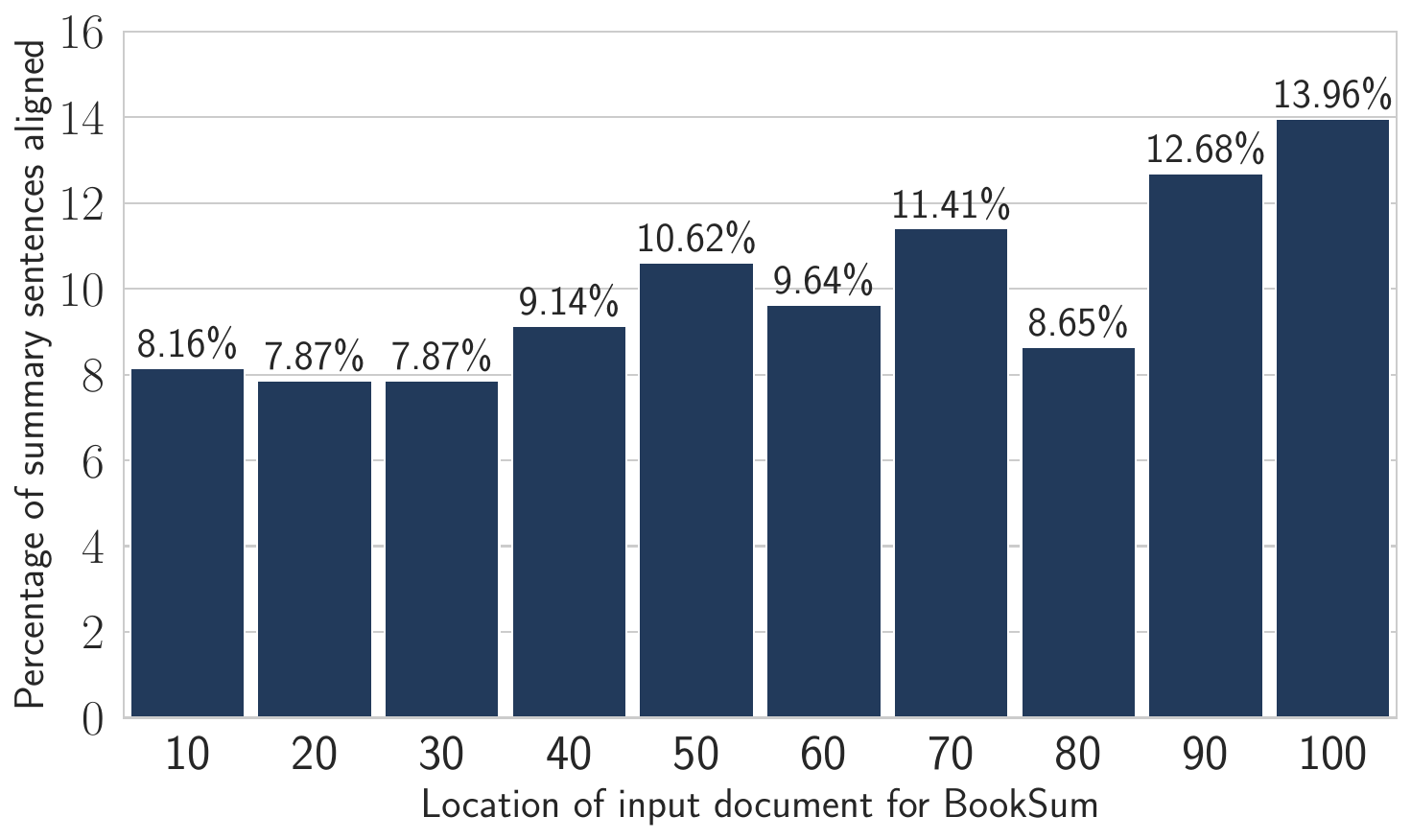}
  \caption{Percentage of generated summary sentences aligned with different segments (bins) of a book in the BookSum test set. CachED BART$_{\text{large}}$ model uses the entire document, with the alignment evenly distributed across segments.}
  \label{fig:summary_location}
\end{figure}

\section{Analyses}

In this section, we present a comprehensive analysis of various aspects related to our approach and its performance. For our analysis, we will use the CachED BART$_{\text{large}}$ model, which performs the best in our experiments. We first examine the utilization of the full context to understand how effectively our method handles and processes long inputs. Next, we analyze the method's performance in relation to document length, assessing how changes in length impact the model's efficiency and accuracy. We also address time and memory usage to evaluate the computational resources required by our approach. Finally, we assess the factual consistency of the model's outputs, ensuring that the generated summaries are not only better in terms of ROUGE but are also factually accurate.

\begin{figure}
  \centering
  \includegraphics[width=\columnwidth]{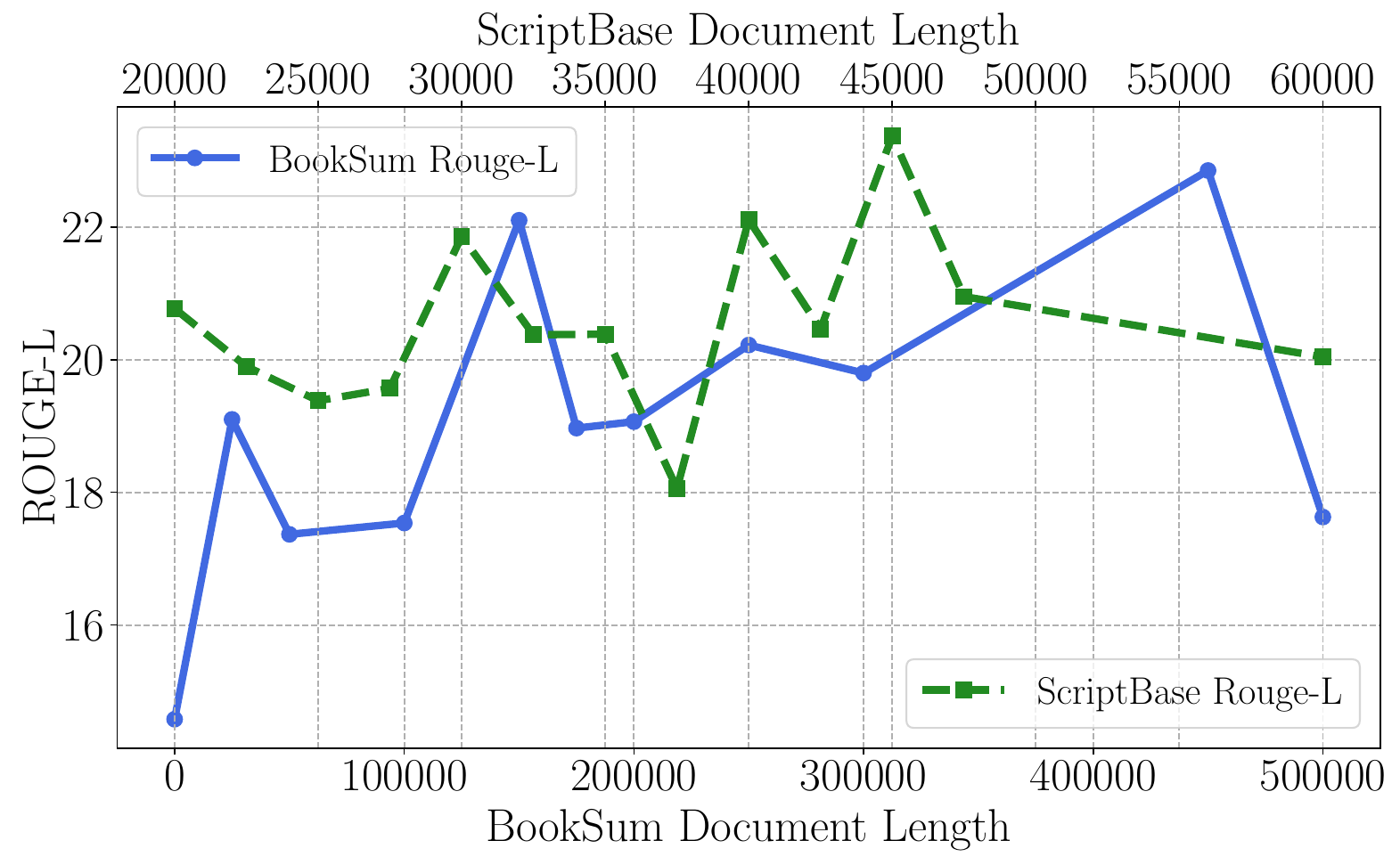}
  \caption{ROUGE-L of the summaries generated by CachED BART$_{\text{large}}$ model for documents of different lengths.
  The model maintains consistent performance across various document lengths.}
  \label{fig:rouge_length}
\end{figure}

\subsection{Utilization of the entire document}

To evaluate whether CachED models effectively use the full context of a document, we use the alignment between the generated summary and the document as a proxy. We first divide the document into equal segments (bins) and map the summary sentences to these segments to estimate from which part of the document they were generated. We use ROUGE-L-based alignment~\citep{chen-bansal-2018-fast,zhang-etal-2022-summn}, a method previously used for the automatic generation of source-summary pairs. Based on the ROUGE-L scores, we select and map the best segment for each generated summary sentence. This approach allows us to analyze the parts of the document that contribute to the summary.

Figure~\ref{fig:summary_location} presents the alignment statistics, showing where in a document that a summary sentence aligns to in the BookSum test set. Our results indicate that the CachED BART$_{\text{large}}$ model does not exhibit significant position bias and utilizes the document uniformly. Overall, the summary sentences align slightly more to the end of a book compared to the first half of the book.

\subsection{Performance and document length}

To further verify the use of full context, we study how sensitive the performance of the CachED BART$_{\text{large}}$ model is to documents of different lengths.
Figure~\ref{fig:rouge_length} shows ROUGE-L scores of our model on the ScriptBase and BookSum test sets.
The ROUGE-L performance remain relatively consistent across documents of different lengths, indicating that our approach is robust in handling extremely long documents.

In the case of the BookSum dataset, the ROUGE-L scores exhibit a stable trend around the mean of 18.8, despite the document lengths reaching up to 500K tokens. Similarly, for the ScriptBase dataset, the model achieves ROUGE-L scores between 18 and 22, relatively insensitive to the document length.

The consistent performance across varying document lengths shows that the CachED BART$_{\text{large}}$ model is using the full context when summarizing documents.

\begin{figure}
  \centering
  \includegraphics[width=\columnwidth]{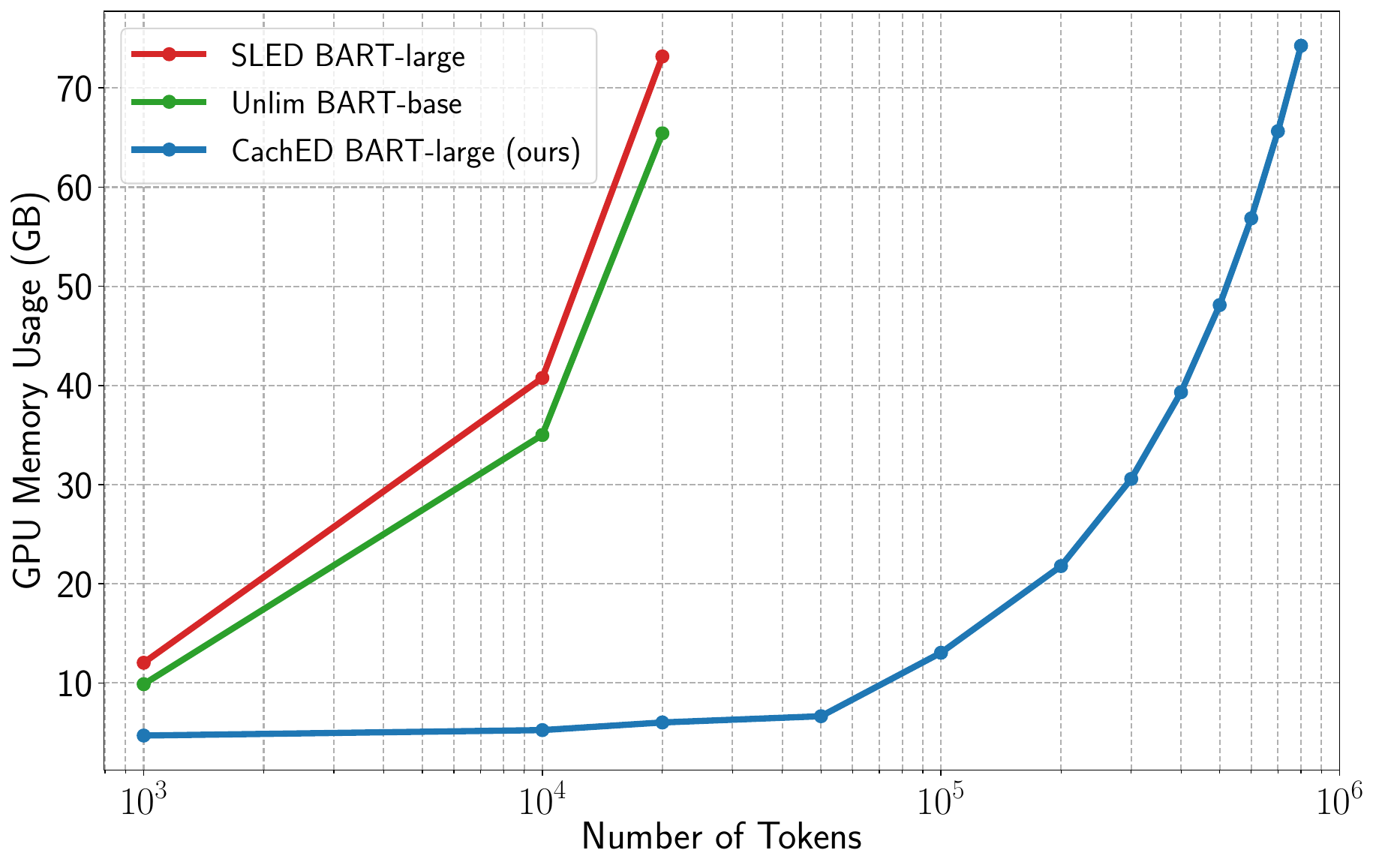}
  \includegraphics[width=\columnwidth]{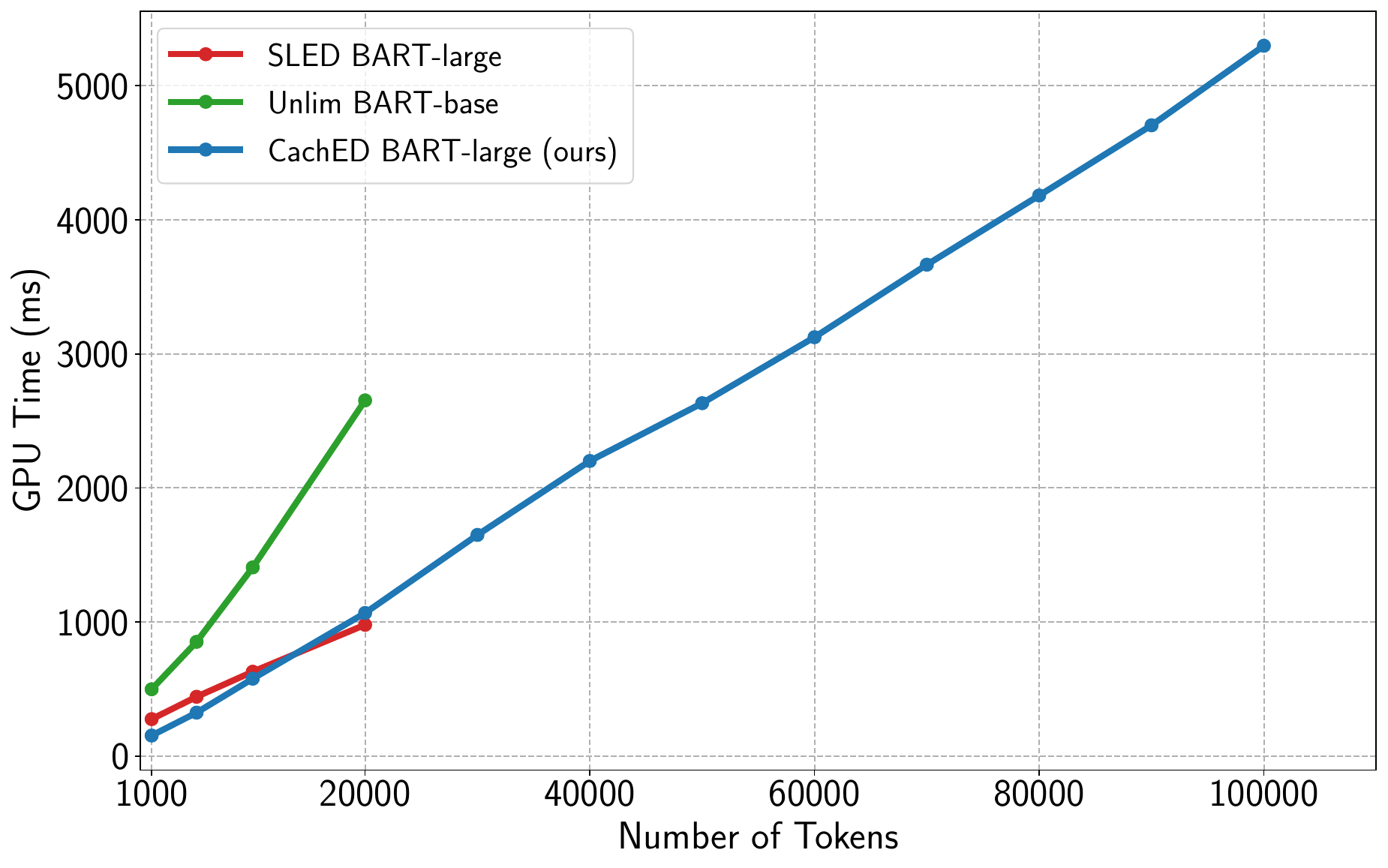}
  \caption{Comparison of GPU memory usage (top) and time usage (bottom) for SLED, Unlimiformer, and CachED BART$_{\text{large}}$. \label{fig:memory_time_usage}}
\end{figure}

\subsection{Time and Memory Usage}
\color{black}
To understand the time and memory consumption, we measure GPU memory and wall-clock time for SLED, Unlimiformer, and our proposed approach across varying numbers of tokens. For a fair comparison, all experiments were performed on a single A100 80GB GPU with the same batch size and full precision. 

Figure~\ref{fig:memory_time_usage} (top) compares GPU memory usage (in GB) as the number of tokens increases. SLED and Unlimiformer show a steep increase in memory consumption as the token count grows. In contrast, CachED BART$_{\text{large}}$ demonstrates more linear memory scaling. Its memory usage remains below 20GB, even when the input reaches $10^5$ tokens. Given a GPU with 80GB of memory, CachED BART$_{\text{large}}$ can compute gradients up to nearly one million tokens, whereas both SLED and Unlimiformer hit memory limitations at much smaller scales (around 20,000 tokens). 

Next, we analyze the impact of feeding forward the encoder twice in our approach. Figure~\ref{fig:memory_time_usage} (bottom) shows the wall-clock time (in milliseconds) required to process varying token lengths. As the input size grows, Unlimiformer shows a sharp increase in GPU time consumption, likely due to its k-nearest neighbor search operation.  In contrast, SLED and CachED maintain a similar linear growth as the token count increases. CachED incurs a slight increase in wall-clock time compared to SLED at around 10,000 tokens, which can be attributed to the recomputation of the encoder during backpropagation.  Despite this minor overhead, the processing time remains within a manageable range, ensuring that the model can process extensive texts without huge time overhead. This result confirms our intuition that feeding forward the encoder does not take up much of the time during training compared to other parts of the computation, and that our approach is efficient in practice.
\color{black}
\begin{table}[ht]
\centering
\scalebox{0.9}{
\begin{tabular}{lcc}
\toprule
\multirow{2}{*}{Method} & \multicolumn{2}{c}{\textbf{ScriptBase}} \\
 & AlignScore & FActScore \\
\midrule
Unlim. (BART$_{\text{base}}$) & 31.39 & 44.00 \\
SLED (BART$_{\text{large}}$) & 32.25 & 41.10 \\
CachED BART$_{\text{large}}$ & \textbf{35.04} & \textbf{51.80} \\
\midrule
\multirow{2}{*}{Method} & \multicolumn{2}{c}{\textbf{BookSum}} \\
 & AlignScore & FActScore \\
\midrule
Unlim. (BART$_{\text{base}}$) & 35.00 & 39.30 \\
SLED (BART$_{\text{large}}$) & 36.92 & 37.70 \\
CachED BART$_{\text{large}}$ & \textbf{41.33} & \textbf{52.90} \\
\bottomrule
\end{tabular}
    }
\caption{Results of automatic evaluation of factual consistency on generated summaries for ScriptBase and BookSum dataset.}
\label{tab:factuality_result}
\end{table}
\subsection{Evaluation of Factual Consistency}

To evaluate the performance of our method in generating factually correct summaries, we compute the AlignScore~\citep{zha-etal-2023-alignscore} and FActScore~\citep{min-etal-2023-factscore} on Unlimiformer, SLED, and our CachED BART. AlignScore measures factual consistency based on unified information alignment between the context (input document) and claims (summary sentences). FActScore parses the generated summary into atomic facts and determines whether these facts are supported by a knowledge source. We use the variant proposed by \cite{zha-etal-2023-alignscore}, which uses the gold reference summary as the knowledge source for summary evaluation. We use the GPT-3.5-Turbo model for FActScore and AlignScore-large for AlignScore.

Table~\ref{tab:factuality_result} shows the results for both metrics on summaries generated for the more challenging extremely long document summarization, i.e., on ScriptBase and BookSum. We compare the results of CachED BART$_{\text{large}}$ to Unlimiformer and SLED. Our model generated more factually consistent summaries and substantially outperformed the other models in terms of both the unified alignment of AlignScore and the number of summary facts supported by the gold reference summary measured by FActScore.  Additionally, Table~\ref{tab:summ_length_statistics} in Appendix \ref{apn:summ_len} provides the mean and median lengths of the gold and generated summaries, highlighting that our model produces concise summaries without compromising on factual accuracy. 
\begin{table}[ht]
\vspace{1em}
\centering
\scalebox{0.9}{
\begin{tabular}{lcccc}
\toprule
\multicolumn{5}{c}{\textbf{ScriptBase}} \\
\midrule
\textbf{Method} & \textbf{R1} & \textbf{R-2} & \textbf{R-L} & \textbf{BS-F1} \\
\midrule
Llama 3.1 8B (ZS)       & 14.63 & 2.17  & 13.68 & 43.97 \\
Llama 3.1 8B (FT)  & 23.34 & 4.8 & 18.63 &46.98 \\
GPT-4o (ZS)            & 42.02 & 10.2  & \textbf{38.94} & 56.24 \\
CachED BART$_\text{large}$ & \textbf{50.2} & \textbf{14.9} & 20.4 & \textbf{64.6} \\
\midrule
\multicolumn{5}{c}{\textbf{BookSum}} \\
\midrule
Llama 3.1 8B (ZS)       & 12.61 & 2.12  & 11.81 & 41.28 \\
Llama 3.1 8B (FT)  & 29.19 & 4.17 & \textbf{28.06} & 50.12\\
GPT-4o (ZS)            & 29.97 & 6.44  & 27.92 & 51.62 \\
CachED BART$_\text{large}$ & \textbf{42.8} & \textbf{10.5} & 18.8 & \textbf{54.4} \\
\bottomrule
\end{tabular}
}
 \caption{Comparison of our method with GPT-4o and Llama 3.1 8B on extremely long document summarization datasets.}
\label{tab:llm_comparison}
\end{table}
\subsection{Comparison with Large Language Models}

Table~\ref{tab:llm_comparison} compares the performance of our approach, CachED BART$_\text{large}$, with the large language models GPT‑4o (zero‑shot) \citep{openai2024gpt4ocard} and Llama 3.1 8B \citep{grattafiori2024llama3herdmodels} under both zero‑shot and Low-rank Adaptation~\citep[LoRA;][]{LoRA} Fine‑Tuning (FT) settings on two extremely long–document summarization datasets: ScriptBase and BookSum. All models were evaluated with a context length of 128K tokens during inference. Please refer to Appendix~\ref{apn:hparams} for fine‑tuning hyperparameters and Appendix~\ref{apn:prompt_template} for prompt templates.

On ScriptBase, LoRA fine‑tuning improves Llama 3.1 8B over its zero‑shot baseline across all metrics. GPT‑4o (ZS) performs substantially better than Llama 3.1 8B (FT) and achieves the best ROUGE‑L (38.94). For ROUGE‑1, ROUGE‑2, and BERTScore, however, CachED BART$_\text{large}$ outperforms the LLM variants.  

For the BookSum dataset, LoRA‐tuned Llama 3.1 8B reaches performance close to zero-shot GPT-4o and achieves an overall better ROUGE-L score of 28.06. CachED BART$_\text{large}$ perform better in terms of ROUGE-1/2 and BERTScore.

We observed that these LLMs occasionally exhibit memorization on our test sets, yet frequently produce incoherent or repetitive passages drawn verbatim from the source.  We discuss this behavior in more detail in Appendix~\ref{apn:llm_memorization}.

\begin{figure}
  \centering
  \includegraphics[width=\columnwidth]{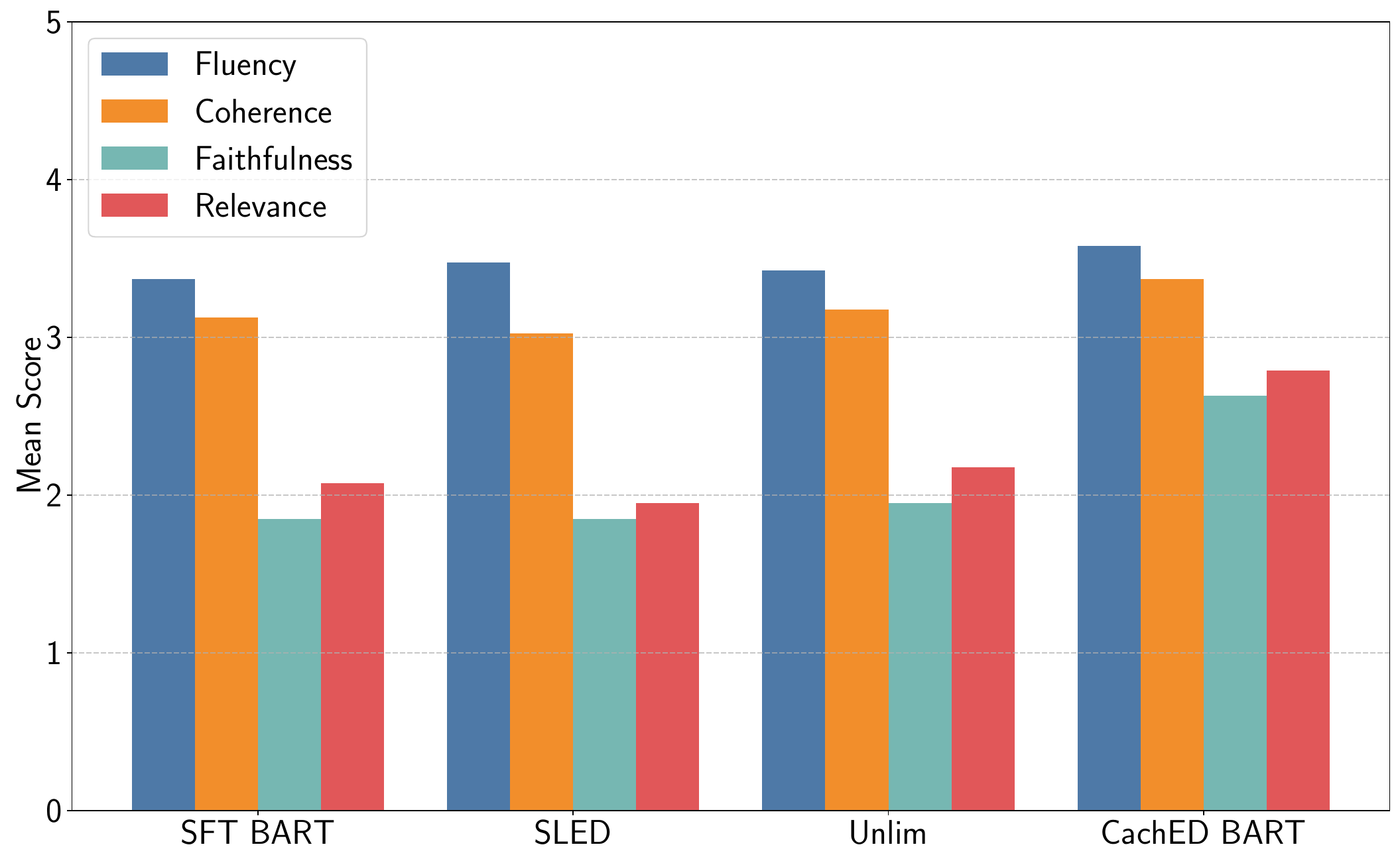}
  \caption{Mean scores for Fluency, Coherence, Faithfulness, and Relevance from the human evaluation of generated summaries on the QMSUM dataset. CachED BART achieves the highest scores for all the dimensions outperforming baseline models (SFT BART, SLED, and Unlimiformer).}
  \label{fig:human_eval}
\end{figure}
\subsection{Human Evaluation}

We conducted a human evaluation to assess the quality of summaries generated by our model in comparison to baseline approaches. A total of 40 crowdworkers were recruited through Prolific, meeting the following criteria: L1 English speakers, holders of an undergraduate degree, and a minimum of 100 previously approved studies. Participants were compensated at a rate of \$17 per hour.

The evaluation focused on four key dimensions: \textbf{Fluency}, \textbf{Coherence}, \textbf{Faithfulness}, and \textbf{Relevance}, similar to \citet{10.1162/tacl_a_00373}. Each participant rated summaries on a Likert scale ranging from 1 (Poor) to 5 (Excellent) for each dimension. The evaluation was conducted on a sample of 20 summaries from the QMSUM dataset.  We selected QMSUM because its reference and generated summaries are relatively shorter than those of other datasets, making it more practical for a focused human evaluation. This allowed for direct comparison of summaries from different models while reducing the cost and complexity of the evaluation process. Full instructions for the evaluation are provided in Appendix~\ref{apn:human_eval_appen}. We compared the performance of CachED BART against the top-performing variant of SFT BART, SLED, and Unlimiformer.

\paragraph{Result} The mean scores for each metric across the four models are shown in Figure~\ref{fig:human_eval}.
 All models performed comparably in Fluency and Coherence with CachED BART achieving slightly higher scores, with means of 3.6 and 3.4, respectively. CachED BART achieved the highest mean scores for Faithfulness (2.6) and Relevance (2.8), outperforming SFT BART, SLED, and Unlimiformer.
 
\paragraph{Statistical Analysis} We performed a one-way ANOVA to assess whether there were statistically significant differences among the models across the four dimensions.
The results showed no significant differences for Fluency ($F=0.305$, $p=0.822$) and Coherence ($F=0.634$, $p=0.594$), indicating that model performance did not differ on these metrics. In contrast, significant differences were observed for Faithfulness ($F=4.875$, $p=0.0029$) and Relevance ($F=4.242$, $p=0.0065$).

To identify the specific differences among the models, Tukey's HSD post-hoc tests were performed for the significant dimensions. For Faithfulness, CachED BART significantly outperformed SFT BART ($p=0.0075$), SLED ($p=0.0075$), and Unlimiformer ($p=0.0261$). Similarly, for Relevance, CachED BART was significantly better than SFT BART ($p=0.0288$) and SLED ($p=0.0066$), while other comparisons did not yield significant differences. These results suggest that CachED BART generates more faithful and relevant summaries compared to the baseline models.

\section{Gradient Caching for Decoder‑Only Models}
\label{sec:decoder_only}

While this work focuses on encoder–decoder architectures, the gradient caching framework can be extended to decoder‑only transformer models by using the same chunk and recompute paradigm. Concretely, given an input sequence of length $L$, we split it into $K$ non‑overlapping chunks of size similar to parallel-context processing used in long-context LLMs~\citep{grattafiori2024llama3herdmodels}.
During the forward pass, each chunk is processed sequentially, and the resulting key/value (KV) tensors are detached and cached. After computing the loss over the concatenated chunk outputs, a second pass then replays each chunk’s forward pass and propagates the gradients to the respective blocks.  

Moreover, our gradient‑caching method can also be applied with ring/block attention~\citep{ringattention}, where chunk processing can be done on a single GPU in a loop by removing the computation graph and caching the gradients and then relaying the gradient to complete backpropagation. 

We leave a thorough empirical study of decoder‑only gradient caching to future work, including benchmarking models and evaluating the trade‑offs between extra forward computation, memory savings, and end‑to‑end performance on long summarization tasks. 

\section{Related Work}
\subsection{Efficient Transformers}

Prior work has investigated reducing the quadratic complexity of self-attention through efficient attention mechanisms. BigBird~\citep{bigbird}, Longformer~\citep{Beltagy2020Longformer}, LongT5~\citep{guo-etal-2022-longt5}, and ETC~\citep{ainslie-etal-2020-etc} utilize sparse attention by restricting attention to a set of local and global tokens, thereby enabling the processing of long documents. Additionally, Linformer~\citep{wang2020linformerselfattentionlinearcomplexity} computes self-attention using a low-rank matrix. Routing transformer~\citep{routing} applies a sparse routing module based on online k-means to self-attention, reducing the overall complexity. Performer~\citep{performer} employs a kernel-based estimation of attention, while Reformer~\citep{Kitaev2020Reformer} uses locality-sensitive hashing to reduce attention complexity. These methods typically require pretraining from scratch instead of being directly integrated into existing pre-trained models. Recently, \citet{han-etal-2024-lm-Infinite} proposed an attention mask for zero-shot length generalization of large language models (LLMs), which extends context only at inference time.

\subsection{Parallel Encoder/Chunk Processing}

Previous work has also attempted to overcome the limitation of processing long context lengths by dividing the input into chunks and processing each chunk individually. SLED~\citep{ivgi-etal-2023-efficient} splits the long sequence into overlapping chunks and processes each chunk with the encoder, then fuses it in the decoder. Unlimiformer~\citep{unlimiformer} extends SLED by offloading the cross-attention to k-nearest neighbors ($k$NN) indexing. Both approaches are similar to our work as they can be applied to any pre-trained encoder-decoder model without additional parameters; however, they truncate the input length during training to 16K tokens and do not utilize the full context. PageSum~\citep{liu-etal-2022-leveraging-locality} performs encoding and decoding separately for individual chunks, with the final outputs being a weighted combination of local predictions. This method adds new parameters for weighting, and the generated tokens have strict locality due to independent decoding. \citet{ratner-etal-2023-parallel} employ parallel context windows for LLMs to extend context length during inference, improving in-context learning, but they do not extend the context length of the models during training.

More recently, \citet{xie2024chunkalignselectsimple} propose parallel chunking with reinforcement learning-based selection, adding parameters to the model and truncating the input during training to 16K tokens. \citet{yen2024longcontextlanguagemodelingparallel} proposed extending the context length of decoder-only models with parallel encoding by freezing the decoder layer and adding new cross-attention layers. Our approach does not add any additional parameters to the model and can perform full fine-tuning.

\subsection{Long Context Modeling}

Recent work on LLMs has focused on extending the context length of the models. \citet{ratner-etal-2023-parallel} employ parallel context windows for LLMs to extend context length during inference, improving in-context learning. However, they do not extend the context length of the models during training. Other research includes extrapolating positional embeddings to extend the context window without fine-tuning~\citep{chen2023extendingcontextwindowlarge,peng2024yarn}. Additionally, \citet{xiao2024efficient} have proposed window-based attention during inference. Our approach aims to utilize the full context during training. Recently, \citet{munkhdalai2024leavecontextbehindefficient} propose infini-attention by incorporating compressive memory into the attention mechanism. This method adds new parameters to the model and requires continual pretraining.

\section{Conclusion}

In this work, we propose CachED, a novel approach for enabling end-to-end training of existing encoder-decoder models for extremely long document summarization by leveraging gradient caching. Our approach efficiently processes long input sequences without truncation, allowing for full-context utilization during both training and inference. The experimental results show that our approach surpasses existing approaches across multiple datasets. We show substantial improvements in ROUGE and BERTScore, even with a smaller BART$_{\text{base}}$ model, highlighting the effectiveness of our method. Moreover, our approach does not add additional parameters to the model, maintaining the original model's architecture while enhancing its capability to handle long documents. 
Future research can explore the applicability of our framework to a broader range of models, including decoder-only large language models, and further improve chunk processing strategies to capture more global context information.

\section{Limitation}

One limitation of our work is that we have only focused on encoder-decoder models. We hope that future work can investigate the applicability of our method to decoder-only large language models. Additionally, we only apply our approach to BART, but we look forward to using it with other encoder-decoder models.

Another limitation is that our method encodes chunks independently, restricting self-attention to local contexts within each chunk. This can potentially reduce the model's ability to capture long-range dependencies across chunks. Future work could explore the use of overlapping chunks or additional attention layers over the chunk representations to mitigate this issue and enhance the model's ability to capture wider context.

\section*{Acknowledgements}
We thank the action editor and anonymous reviewers for their constructive feedback. This work was supported in part by the UKRI Centre for Doctoral Training in Natural Language Processing, funded by the UKRI (grant EP/S022481/1) and the School of Informatics at the University of Edinburgh.

\bibliography{tacl2021}
\bibliographystyle{acl_natbib}

\iftaclpubformat
\onecolumn
\fi
\appendix
\section{Implementation Details}\label{apn:hparams}

\begin{table}[ht]
    \centering
\small
    \begin{tabular}{@{~}l@{~}|@{~}l@{~}}
    \textbf{Parameter} & \textbf{Value} \\
    \hline\hline
    Learning rate & 1e-5 \\
    Optimizer & AdamW~\citep{adam} \\
    AdamW $\beta_1$ & 0.9\\
    AdamW $\beta_2$ & 0.99\\
    batch size & 1 \\
    Effective batch size & 2 \\
    Warmup Strategy & linear \\ 
    Warmup Steps & 1024 \\ 
    Decoding Strategy & Greedy \\
    Sample & False \\
    Beam size & 4\\
    Chunk Size (BART) & 1024 \\
    Chunk Size (T5) & 512 \\
    \hline
    \multicolumn{2}{l}{LoRA FT}\\
    \hline
    rank & 16 \\
    lora\_alpha & 16 \\
    lora\_dropout & 0.05 \\
    max\_seq\_length & 10240 \\
    epochs & 15 \\
    target\_modules & q\_proj, k\_proj, v\_proj,\\
    & o\_proj, gate\_proj, up\_proj,\\
    & down\_proj    
    \vspace{1mm}
    \end{tabular}
    \caption{Hyperparameter values used for our experiments.}
    \label{tab:hyperparams}
\end{table}

All experiments were performed on an single A100 GPU with 80GB memory except LoRA FT which were performed on 2 A100s with 80GB memory. GovReport was trained for 10 epochs, QMSUM for 20 epochs, and ScriptBase, BookSum, and SummscreenFD were trained for 15 epochs. For evaluation, we used ROUGE metric Perl package \footnote{\href{https://github.com/summanlp/evaluation/tree/master/ROUGE-RELEASE-1.5.5}{https://github.com/summanlp/evaluation/tree/
master/ROUGE-RELEASE-1.5.5}} and BERTScore with the microsoft/deberta-xlarge-mnli model.

\section{Instructions for Human Evaluation}\label{apn:human_eval_appen}

In this task, you will assess the quality of computer-generated summaries by comparing them against gold (reference) summaries.
You will be provided with a \textbf{Gold Summary} and a corresponding \textbf{Generated Summary}.
\\
Your task is to evaluate the generated summary on the following four dimensions: \textbf{Fluency}, \textbf{Coherence}, \textbf{Relevance}, and \textbf{Faithfulness}.
\\
You will assign a score between \textbf{1 (Poor)} and \textbf{5 (Excellent)} for each dimension.
\textbf{Dimensions of Evaluation:}
\paragraph{1. Fluency:} This dimension evaluates whether the generated summary is grammatically correct, easy to read, and well-structured.
\paragraph{2. Coherence:} This dimension assesses whether the sentences in the generated summary flow logically and maintain a consistent narrative.
\paragraph{3. Faithfulness:} This dimension checks if all the facts presented in the generated summary are accurate and can be directly inferred from the gold summary.
\paragraph{4. Relevance:} This dimension evaluates whether all the important facts from the gold summary are present in the generated summary.

\paragraph{Task Instructions:}
For each gold summary and the corresponding generated summary:
\begin{enumerate}
  \item Rate the generated summary for each dimension (Fluency, Coherence, Faithfulness, Relevance) on a scale of 1 to 5 based on the definitions provided.
  \item Ignore minor grammatical errors or abrupt endings that do not significantly impact the overall quality of the summary.
  \item Make your selection based solely on the definitions and examples provided for each dimension.
\end{enumerate}

\section{Prompt Template}\label{apn:prompt_template}
The prompt templates used in our experiments.
\\\\
\textbf{Prompt Template for LLaMA 3.1 8B}
\begin{lstlisting}
<|begin_of_text|><|start_header_id|>system<|end_header_id|>You are a helpful assistant<|eot_id|><|start_header_id|>user<|end_header_id|>Summarize the following text in detail like a plot summary and include all the main events.\n[Input Text]<|eot_id|><|start_header_id|>assistant<|end_header_id|>
\end{lstlisting}
\textbf{Prompt Template for GPT-4}
\begin{lstlisting}
Summarize the following text in detail like a plot summary and include all the main events.\n[Input Text]
\end{lstlisting}

\section{Memorization of Dataset in LLMs}\label{apn:llm_memorization}

\begin{table}[ht]
\vspace{1em}
\centering
\scalebox{0.85}{
\begin{tabular}{lcccc}
\toprule
\textbf{Method} & \textbf{R1} & \textbf{R-2} & \textbf{R-L} & \textbf{BS-F1} \\
\midrule
\multicolumn{5}{c}{\textbf{ScriptBase (Title + Year)}} \\
\midrule
Llama 3.1 8B       & 25.36 & 4.06  & 24.35 & 44.93 \\
GPT-4o           & 45.03 & 11.68  & 41.85 & 57.36 \\
\midrule
\multicolumn{5}{c}{\textbf{ScriptBase (Zero-Shot)}} \\
\midrule
Llama 3.1 8B        & 14.63 & 2.17  & 13.68 & 43.97 \\
GPT-4o           & 42.02 & 10.2  & 38.94 & 56.24 \\
\midrule
\multicolumn{5}{c}{\textbf{BookSum (Title + Year)}} \\
\midrule
Llama 3.1 8B        & 32.67 & 4.83  & 31.71 & 56.00 \\
GPT-4o             & 38.65 & 9.69  & 36.31 & 55.75 \\
\midrule
\multicolumn{5}{c}{\textbf{BookSum (Zero-Shot)}} \\
\midrule
Llama 3.1 8B (ZS)       & 12.61 & 2.12  & 11.81 & 41.28 \\
GPT-4o (ZS)            & 29.97 & 6.44  & 27.92 & 51.62 \\
\bottomrule
\end{tabular}
}
 \caption{Performance of Llama 3.1 8B and GPT-4o on summarization when prompted with title+year only vs. full document. Higher performance with title+year indicates possible memorization.}
\label{tab:llm_memorization}
\end{table}
To investigate whether large language models rely on memorized summaries, we compare two prompt settings on ScriptBase and BookSum:
\begin{itemize}
  \item \textbf{Title + Year only}: the model is prompted with only the title and release/publication year of the movie or book (no document text).
  \item \textbf{Full context (Zero‑Shot)}: the model is given the entire document as input.
\end{itemize}
Table~\ref{tab:llm_memorization} shows that both Llama 3.1 8B and GPT‑4o achieve substantially higher ROUGE and BERTScore when prompted with only the title and year, compared to using the full text. On ScriptBase, Llama 3.1 8B’s ROUGE‑L increases from 13.68 (full context) to 24.35 (title + year), and GPT‑4o improves from 38.94 to 41.85. Similarly, on BookSum, Llama 3.1 8B's ROUGE-L improves from 11.81 to 31.71, and GPT‑4o's from 27.92 to 36.31.

These results suggest that both models likely rely on memorized content from pretraining rather than processing long inputs effectively. The performance drop when given the complete document indicates a potential limitation in their ability to handle extremely long sequences effectively.

\section{Summary length statistics}\label{apn:summ_len}

Table~\ref{tab:summ_length_statistics} reports the mean and median token lengths of the gold (reference) summaries alongside those generated by each model on the ScriptBase and BookSum test sets.

\begin{table}[ht]
\vspace{1em}
\centering
\scalebox{0.9}{
\begin{tabular}{lcc}
\toprule
\multicolumn{3}{c}{\textbf{ScriptBase}} \\
\midrule
\textbf{Method} & \textbf{Mean} & \textbf{Median}\\
\midrule
Gold Summary      & 878.32 & 906.5 \\
Unlim. (BART$_{\text{base}}$) & 856.72 &851.0\\
SLED (BART$_{\text{large}}$) & 847.54 & 848.5\\
CachED BART$_\text{large}$ & 844.2 & 887.0\\
Llama 3.1 8B (ZS) & 904.64 & 1026.5 \\
Llama 3.1 8B (FT) & 1433.04  & 1029.0 \\
GPT-4o & 714.16  & 704.0\\
\midrule
\multicolumn{3}{c}{\textbf{BookSum}} \\
\midrule
Gold Summary      & 859.39 & 1024.0 \\
Unlim. (BART$_{\text{base}}$) & 964.7 & 1022.0\\
SLED (BART$_{\text{large}}$) & 995.04 & 1017.0\\
CachED BART$_\text{large}$ & 847.89 & 927.0\\
Llama 3.1 8B (ZS) & 1037.0 & 1017.0\\
Llama 3.1 8B (FT) & 979.65 & 1029.0\\
GPT-4o & 734.20  & 719.5\\
\bottomrule
\end{tabular}
}
\caption{Mean and median lengths (in tokens) of gold and generated summaries across the ScriptBase and BookSum datasets.}
\label{tab:summ_length_statistics}
\end{table}

\section{Sample of Book Summaries}\label{sec:generatedsumm}
We present sample summaries of the book \textbf{Sense and Sensibility} generated using SLED, Unlimiformer, and our approach CachED BART.
\subsection{Gold Summary}
The Dashwood family is introduced; Mr. and Mrs. Dashwood and their three daughters live at Norland Park, an estate in Sussex. Unfortunately, Mr. Dashwood's wife and daughters are left with very little when he dies and the estate goes to his son, John Dashwood. John and his wife Fanny have a great deal of money, yet refuse to help his half-sisters and their mother. Elinor, one of the Dashwood girls, is entirely sensible and prudent; her sister, Marianne, is very emotional and never moderate. Margaret, the youngest sister, is young and good-natured. Mrs. Dashwood and her daughters stay at Norland for a few months, mostly because of the promising friendship developing between Elinor and Edward Ferrars, Fanny's shy, but very kind, brother. Elinor likes Edward, but is not convinced her feelings are mutual; Fanny is especially displeased by their apparent regard, as Edward's mother wants him to marry very well. A relative of Mrs. Dashwood's, Sir John Middleton, offers them a cottage at Barton Park in Devonshire; the family must accept, and are sad at leaving their home and having to separate Edward and Elinor. They find Barton Cottage and the countryside around it charming, and Sir John Middleton a very kind and obliging host. His wife, Lady Middleton, is cold and passionless; still, they accept frequent invitations to dinners and parties at Barton Park. The Dashwoods meet Mrs. Jennings, Sir John's mother-in-law, a merry, somewhat vulgar older woman, and Colonel Brandon, a gentleman and a bachelor. The Colonel is soon taken with Marianne, but Marianne objects to Mrs. Jennings attempts to get them together, and to the "advanced" age and serious demeanor of the Colonel. Marianne falls and twists her ankle while walking; she is lucky enough to be found and carried home by a dashing man named Willoughby. Marianne and Willoughby have a similar romantic temperament, and Marianne is much pleased to find that Willoughby has a passion for art, poetry, and music. Willoughby and Marianne's attachment develops steadily, though Elinor believes that they should be more restrained in showing their regard publicly. One pleasant day, the Middletons, the Dashwoods, and Willoughby are supposed to go on a picnic with the Colonel, but their plans are ditched when Colonel Brandon is forced to leave because of distressing news. Willoughby becomes an even more attentive guest at the cottage, spending a great deal more time there than Allenham with his aunt. Willoughby openly confesses his affections for Marianne and for all of them, and hopes they will always think of him as fondly as he does of them; this leaves Mrs. Dashwood and Elinor convinced that if Marianne and Willoughby are not engaged, they soon will be. One morning, Mrs. Dashwood, Elinor, and Margaret leave the couple, hoping for a proposal; when they return, they find Marianne crying, and Willoughby saying that he must immediately go to London. Mrs. Dashwood and Elinor are completely unsettled by this hasty departure, and Elinor fears that they might have had a falling-out. Marianne is torn up by Willoughby's departure, and Elinor begins to question whether Willoughby's intentions were honorable. But, whether Willoughby and Marianne are engaged remains a mystery, as Marianne will not speak of it. Edward comes to visit them at Barton, and is welcomed very warmly as their guest. It is soon apparent that Edward is unhappy, and doesn't show as much affection for Elinor; when they spot a ring he is wearing, with a lock of hair suspiciously similar to Elinor's, even Elinor is baffled. Edward finally forces himself to leave, still seeming distressed. Sir John and Mrs. Jennings soon introduce Mrs. Jennings' other daughter, Mrs. Palmer, and her husband to the family. Mrs. Palmer says that people in town believe that Willoughby and Marianne will soon be married, which puzzles Elinor, as she knows of no such arrangements herself. Elinor and Marianne meet the Middletons' new guests, the Miss Steeles, apparently cousins; they find Miss Steele to be nothing remarkable, while Lucy is very pretty but not much better company. However, the Miss Steeles instantly gain Lady Middleton's admiration by paying endless attention to her obnoxious children. Elinor, unfortunately, becomes the preferred companion of Lucy. Lucy inquires of Mrs. Ferrars, which prompts Elinor to ask about her acquaintance with the Ferrars family; Lucy then reveals that she is secretly engaged to Edward. It turns out that Edward and Lucy knew each other while Edward studied with Lucy's uncle, Mr. Pratt, and have been engaged for some years. Although Elinor is first angry about Edward's secrecy, she soon sees that marrying Lucy will be punishment enough, as she is unpolished, manipulative, and jealous of Edward's high regard for Elinor. The Miss Steeles end up staying at Barton Park for two months. Mrs. Jennings invites Marianne and Elinor to spend the winter with her in London. Marianne is determined to go to see Willoughby, and Elinor decides she must go too, because Marianne needs Elinor's polite guidance. They accept the invitation, and leave in January. Once in town, they find Mrs. Jennings' house comfortable, and their company less than ideal; still, they try their best to enjoy it all. Marianne anxiously awaits Willoughby's arrival, while Elinor finds her greatest enjoyment in Colonel Brandon's daily visits. Elinor is much disturbed when Colonel Brandon tells her that the engagement between Marianne and Willoughby is widely known throughout town. At a party, Elinor and Marianne see Willoughby; Marianne approaches him, although he avoids Marianne, and his behavior is insulting. Marianne angrily writes Willoughby, and receives a reply in which he denies having loved Marianne, and says he hopes he didn't lead her on. Marianne is deeply grieved at being deceived and dumped so coldly; Elinor feels only anger at Willoughby's unpardonable behavior. Marianne then reveals that she and Willoughby were never engaged, and Elinor observes that Marianne should have been more prudent in her affections. Apparently, Willoughby is to marry the wealthy Lady Grey due to his constant need for money. Colonel Brandon calls after hearing the news, and offers up his knowledge of Willoughby's character to Elinor. Colonel Brandon was once in love with a ward to his family, Eliza, who became a fallen woman and had an illegitimate daughter. Colonel Brandon placed the daughter, Miss Williams, in care after her mother's death. The Colonel learned on the day of the Delaford picnic that she had become pregnant, and was abandoned by Willoughby. Elinor is shocked, though the Colonel sincerely hopes that this will help Marianne feel better about losing Willoughby, since he was not of solid character. The story convinces Marianne of Willoughby's guilt, though it does not ease her mind. Out of sympathy, Marianne also stops avoiding the Colonel's company and becomes more civil to him. Willoughby is soon married, which Marianne is grieved to hear; then, again unfortunately, the Miss Steeles come to stay with the Middletons. John and Fanny Dashwood arrive, and are introduced to Mrs. Jennings, and to Sir John and Lady Middleton, deeming them worthy company. John reveals to Elinor that Edward is soon to be married to Miss Morton, an orphan with a great deal of money left to her, as per the plans of his mother. At a dinner party given by John and Fanny for their new acquaintance, Mrs. Ferrars is present, along with the entire Barton party. Mrs. Ferrars turns out to be sallow, unpleasant, and uncivil; she slights Elinor, which hurts Marianne deeply, as she is Edward's mother. The Miss Steeles are invited to stay with John and Fanny. But, Mrs. Jennings soon informs them that Miss Steele told Fanny of Lucy and Edward's engagement, and that the Ferrars family threw the Steele girls out in a rage. Marianne is much grieved to hear of the engagement, and cannot believe that Elinor has also kept her knowledge of it a secret for so long. Edward is to be disinherited if he chooses to marry Lucy; unfortunately, Edward is too honorable to reject Lucy, even if he no longer loves her. Financial obstacles to their marriage remain; he must find a position in the church that pays enough to allow them to marry. Much to Elinor's chagrin, the Colonel, although he barely knows Edward, generously offers the small parish at Delaford to him. Elinor is to convey the offer to Edward, though she regrets that it might help the marriage. Edward is surprised at the generous offer, since he hardly knows the Colonel. Edward decides to accept the position; they say goodbye, as Elinor is to leave town soon. Much to Elinor's surprise, Robert Ferrars, Edward's selfish, vain, and rather dim brother, is now to marry Miss Morton; he has also received Edward's inheritance and money, and doesn't care about Edward's grim situation. It is April, and the Dashwood girls, the Palmers, and Mrs. Jennings, and Colonel Brandon set out for Cleveland, the Palmer's estate. Marianne is still feeling grief over Willoughby; she soon becomes ill after her walks in the rain, and gets a serious fever. The Palmers leave with her child; Mrs. Jennings, though, helps Elinor nurse Marianne, and insists that Colonel Brandon stay, since he is anxious about Marianne's health. Colonel Brandon soon sets off to get Mrs. Dashwood from Barton when Marianne's illness worsens. At last, Marianne's state improves, right in time for her mother and the Colonel's arrival; but Willoughby makes an unexpected visit. Elinor is horrified at seeing him; he has come to inquire after Marianne's health and to explain his past actions. Willoughby says he led Marianne on at first out of vanity; he finally began to love her as well, and would have proposed to her, if not for the money. By saying that he also has no regard for his wife, and still loves Marianne, he attempts to gain Elinor's compassion; Elinor's opinion of him is somewhat improved in being assured of his regard for Marianne. Elinor cannot think him a total blackguard since he has been punished for his mistakes, and tells him so; Willoughby leaves with this assurance, lamenting that Marianne is lost to him forever. Mrs. Dashwood finally arrives, and Elinor assures her that Marianne is out of danger; both Mrs. Dashwood and the Colonel are relieved. Mrs. Dashwood tells Elinor that the Colonel had confessed his love for Marianne during the journey from Barton; Mrs. Dashwood wishes the Colonel and Marianne to be married. Elinor wishes the Colonel well in securing Marianne's affections, but is more pessimistic regarding Marianne's ability to accept the Colonel after disliking him for so long. Marianne makes a quick recovery, thanking Colonel Brandon for his help and acting friendly toward him. Marianne finally seems calm and happy as they leave for Barton, which Elinor believes to signal Marianne's recovery from Willoughby. She is also far more mature, keeping herself busy and refusing to let herself languish in her grief. When Marianne decides to talk about Willoughby, Elinor takes the opportunity to tell her what Willoughby had said at Cleveland, and Marianne takes this very well. Marianne also laments her selfishness toward Elinor, and her lack of civility toward most of their acquaintance. Marianne finally says that she could not have been happy with Willoughby, after hearing of his cruelty toward Miss Williams, and no longer regrets him. The family is stunned when one of their servants returns with news that Edward is married to Lucy, as he just saw them in the village. Elinor knows now that Edward is lost to her forever. Mrs. Dashwood sees how upset Elinor is, and realizes that Elinor felt more for Edward than she ever revealed. One afternoon, Elinor is convinced that the Colonel has arrived at the cottage, but is surprised to find that it is Edward instead. Their meeting is awkward at best; he soon informs them that it is his brother who has been married to Lucy, and not him. Elinor immediately runs from the room, crying out of joy; Edward then senses Elinor's regard for him, and proposes to her that afternoon. Elinor accepts and he gains Mrs. Dashwood's consent to the match. Edward admits that any regard he had for Lucy was formed out of idleness and lack of knowledge; he came to regret the engagement soon after it was formed. After leaving London, Edward received a letter from Lucy saying that she had married his brother Robert, and has not seen her since; thus, he was honorably relieved of the engagement. After receiving the letter, he set out for Barton immediately to see Elinor. Edward will still accept the position at Delaford, although he and Elinor again will not have enough money to live on comfortably. The Colonel visits Barton, and he and Edward become good friends. Edward then becomes reconciled with his family, although he does not regain his inheritance from Robert. His mother even gives her consent for his marriage to Elinor, however much she is displeased by it; she gives them ten thousand pounds, the interest of which will allow them to live comfortably. Edward and Elinor are married at Barton that fall. Mrs. Dashwood and her two remaining daughters spend most of their time at Delaford, both to be near Elinor, and out of the hope that Marianne might accept the Colonel. In the two years that have passed, Marianne has become more mature and more grounded; and she does finally change her mind about the Colonel, and accepts his offer of marriage. The Colonel becomes far more cheerful, and soon Marianne grows to love him as much as she ever loved Willoughby. Mrs. Dashwood remains at Barton with Margaret, now fifteen, much to the delight of Sir John, who retains their company. And Elinor and Marianne both live together at Delaford, and remain good friends with each other and each other's husbands.

\subsection{Generated Summary using CachED BART}
Marianne and Elinor Dashwood, two young cousins of the Dashwood family, live with their mother and sister at their cottage in Norland Park, a small country estate in Sussex. The Dashwoods are not wealthy, but they are in good financial circumstances due to the large estate and the money their uncle left them. Marianne and her sister Elinor, rather, Marianne Dashwood are the only children of a wealthy uncle and aunt. Marianne and Elinor are close in age, but Elinors' relationship with Edward Ferrars, the Dashwoods' brother, is more complicated. Edward is the son of a woman who is not wealthy but who has a reputation for being selfish and unworldly. They are in love, but Edward is not interested in them. They are both too young to be married, and they are both unaware of their parents' financial situation. Marianna and Elinor are invited to visit the Dashwars at their home, Barton Park, by their cousin, Sir John Middleton. They spend a pleasant time there, but Marianne is not happy there. She is attracted to Edward, but he is not attracted to her. Marianne's mother, Mrs. Jennings, is not pleased with Marianne's relationship with her son, and she is not impressed by Elinors relationship to Edward. Marianne is also upset that Elinor has been invited to stay at Barton Park. In the meantime, Elinor Dashwood's cousin, John Dashwood has been visiting from London. John is a friend of the family, and is a good-looking young man. He is the brother-in-law to Elinor to Elinsor's sister, Marianne. John and El inor are friends, and have been friends for years. John has been engaged to Elina's sister Lucy, but she has not yet told him she is engaged to Edward Ferrar. John's mother is not at all happy about the match, and John is not sure if he should tell Elinor about it. Marianne does not want to upset her sister, but is happy to hear that Elina has been in London. Elinor and John are friends and neighbors of the Middleton family. John Dashwood and Marianne are friends of the Jenningses, and the Dash wer happy to be friends with them. John, Elinor, and Marianne are all friends of Sir John's, and Elinor is close to them. When John's sister dies, he and Elinor decide to visit her in town. They go to visit John's house, where they are met by Willoughby, Elinor's brother- in-law. Elinor and John Dashwards are friends for a while, and are happy to see each other. John also knows that Willoughby is engaged, but Willoughby is not ready to announce it. Elinor or John Dash waffles on whether to tell Marianne about it, and when Marianne finds out, she is devastated. She decides to keep it a secret from Elinoor and John, and to keep Willough by herself. Marianne becomes ill, and after a few days, Elinor in or out of town with John Dashwalts, Ellinor decides to tell John about Willoughy. John comes to visit, and he and Mariane are very happy to meet Willoughby. It turns out that Willougby is engaged. Marianne tells Elinor that Willohby is married to Lucy, and that she is happy for him. Ellinar is happy that she has been able to keep the secret from John Dashwalds. Marianne and John go to London, and find that Willoughby is a very good-natured, kind, and intelligent young man, and will be a good husband. They become friends, but John is disappointed when he learns that Marianne has been seeing Edward. He does not like Edward, and wants to see Marianne again. Marianne is happy when she learns that Willoughby is in town, and plans to tell El in return. Elinor is happy about this news, and tells John that she will not tell John that Willoyby is marrying Lucy. John does not believe her, and so he does not tell Elinor.

\subsection{Generated Summary using SLED}
The novel begins with the Dashwood family of Norland, England. The Dashwood sisters, Elinor, and Marianne Dashwood, are engaged to a man named Edward Ferrars, a wealthy and wealthy man who lives in the town of Barton Park, near Norland. They are engaged for a short period of time, but they are unable to afford to live there because of their mother's ill health. They decide to move to Barton Park in order to be closer to their family, and they decide to stay there for a few years. They move to Norland to live with their mother-in-law, Mrs. Dashwood and her three daughters, and the Dashwoods decide to live at Barton Park for a while. They begin to fall in love with their new neighbor, Mr. John Dashwood. However, they soon realize that they are not going to be able to afford their new home, and that they will have to move away from Norland for the sake of their children. They also realize that their new house will be too small for them to afford, and so they are forced to move out of their old home. Meanwhile, they decide that they should stay at Norland and move to a new house in Barton Park. They plan to spend their time there, as they have no money to spend on their own. They will move into the new house when they are old enough to afford it, and then they will move to the new home at Barton, where they will live with the family of their older brother, Edward, and his sister, Marianne, who will be living with them for the rest of their lives. In the meantime, they plan to move into a new home in the village of Barton, which will be built by their new neighbors. They hope to find a place to live for their children, but their plans are thwarted when they find out that they cannot afford to move. In order to make this possible, they have to leave Norland with their two daughters, whom they have never met before. They have to decide whether they want to live in Barton or not, and decide whether or not they should move there. When they do decide to do so, they are surprised to learn that they have been given a house in Norland by a wealthy man named Sir John Middleton, a friend of their father's, who is also a wealthy, well-liked man. They make plans to move there, but Sir John decides to stay at Barton for the time being, and he decides to move with his sister and her two daughters. He decides that he will stay with his family in Barton, but he does not want to spend any more time with them than he can afford to spend with them. He also decides to marry a woman of his own, a woman whom he has never seen before, and who he will marry in the future. He is very happy to be with her, and is very fond of her, but his feelings for her are not reciprocated by her. When he is married, he is devastated by the loss of his wife and children, and feels that he cannot live with her. He wants to marry her, so he asks her to marry him, but she refuses. He tells her that she cannot marry him because she is too young, and she cannot afford it. He then decides that she will marry him. He leaves Norland at the end of the year, and when he returns to England, he plans to spend his time at Barton with his sisters. He plans to marry Marianne and their two young daughters, but Marianne refuses to marry them, because she feels that she is not worthy of the money he has given her. She decides to go with him to Barton, and decides to spend her time there with her sister. She and her sister decide to spend some time with him there, and after a few months, she decides to leave the house and move with him. She is very pleased to see that he is happy with the way he is living, and wishes to spend time with her and her daughters. When she returns home, she finds that she has not been able to spend much time with his children, so she is forced to leave them. She then decides to return to her old home, where she has been living with her mother and her sisters, who are all in the same house. She has no money and she is very unhappy with her situation. She does not know what to do with the money she has left, but her mother is very kind and generous. She tells him that she wants to stay with him and that she would be happy with him if he would marry her. After a short time, she and her mother decide to go to Barton to spend a few days with him, where he will spend the summer with his younger sister. They spend a lot of time together, and it is not long before they are reunited. They return to their old house at Barton.

\subsection{Generated Summary using Unlimiformer}
Elinor Dashwood, the eldest daughter of the Dashwood family, is engaged to Edward Ferrars. Elinor and Marianne, who have been engaged to Willoughby for the past four years, are devastated by the news of Edward's engagement to Marianne. The Dashwood sisters, Mrs. and Mrs. Jennings, are also devastated by their sister's news of her sister's engagement with Edward. Marianne and Edward, who are both in love with their younger sister, are sent to Barton Street, where they are invited to meet Colonel Brandon and his wife, Lady Middleton, and the Dashwoods are to visit their sister, Lady Hurst, at their estate at Barton Park. The next morning, Elinore and Elinaor are invited by Colonel Brandon, who is in town to visit his sister, Marianne's sister, and her niece, Elinnor. The Colonel Brandon is very interested in Marianne as well as her sister, but he is not interested in her sister. The following morning, the Colonel Brandon arrives at Barton Street and asks to see Marianne at her house, where he is staying with his sister and his sister's friend, Lady Crawwood, and Mr. Dashwood. When Marianne arrives, she is shocked to see her sister and her sister in the same room. She tells her sister that Marianne has been engaged for three years, and that she has not seen her sister since her engagement to Edward. When she sees Marianne in the garden, she realizes that her sister has been in bed with her brother, and she has no way of knowing what she is going to do with her sister or her niece. She is devastated to find that her brother and sister are in love, but she is not sure what to do about Marianne or her sister; she is also devastated to learn that her younger sister is engaged, to a young man named Colonel Brandon. She does not know what Marianne is doing, but her sister does not want Marianne to know what she has been doing, so she is forced to tell Marianne about her sister-in-law's affair with her, and to try to make Marianne feel better by telling Marianne what she knows. She tries to convince Marianne that she is wrong, but Marianne does not believe her sister is right to know, and so she does not tell her. She decides that she will go to her uncle's house to visit Marianne when Marianne returns to Barton, where she will be staying with her mother and sister. She also decides to ask Marianne if she is interested in seeing her sister at Barton, for she is sure that she would be able to get her sister to marry her brother. She agrees to go to Barton and visit her sister when she returns from Barton, but when she comes back to Barton the next day, she finds that she cannot go with Marianne because she is afraid of Marianne returning to Barton. She knows that she must tell her sister if she ever sees her brother or Marianne again. She goes to her mother's house and asks Marianne for help. She asks Marianna to tell her about the fact that she was engaged to her sister before she left Barton. When her sister comes to her, she tells Marianne the truth about her brother's affair, she feels very guilty about her own feelings for Marianne; and she fears that she may be in danger. She sends Marianne away to Barton to visit her mother, who has recently returned from her wedding to Edward, and then to her sisters and her aunt, Lady Claywood. She and her sisters decide that they must leave Barton and go to London to find a man to marry, and they decide to stay at Barton to see if Marianne can get Marianne out of her feelings for Edward. They decide to leave Barton for a few days, and Ellinor and Elineor decide to return to Barton in order to find Marianne before Marianne leaves. They plan to meet Marianne on her way to Barton when she is ill, but they do not know who will come to see them. They do not have any way of getting Marianne back, so they leave Barton. They go to a nearby cottage, where Marianne visits her sister for a couple of days before she leaves for Barton, and when she sees her sister again, she thinks she is in trouble. She thinks Marianne might be in trouble, and is afraid that she might get her brother to marry Marianne--or Marianne will be in a situation where she is unable to be with her. When they are in Barton, they go to the house of Mr. Hurwood and his brother, Mr. Crawwood. They visit the family at Barton and are surprised to see their sister in a room with a woman who looks like Marianne but is in a state of disinor. They also see a man named Mr. Crawford, whom they have never seen before, and who seems to be in the middle of their situation.
\end{document}